# Language-agnostic, automated assessment of listeners' speech recall using large language models

Björn Herrmann*

ORCID: 0000-0001-6362-3043

Rotman Research Institute,

Baycrest Academy for Research and Education, M6A 2E1, North York, ON, Canada

Department of Psychology,

University of Toronto, M5S 1A1, Toronto, ON, Canada

*Correspondence concerning this article should be addressed to Björn Herrmann, Rotman Research Institute, Baycrest, 3560 Bathurst St, North York, ON, M6A 2E1, Canada. E-mail: bherrmann@research.baycrest.org

# Abstract

Speech-comprehension difficulties are common among older people. Standard speech tests do not fully capture such difficulties because the tests poorly resemble the context-rich, story-like nature of ongoing conversation and are typically available only in a country's dominant/official language (e.g., English), leading to inaccurate scores for native speakers of other languages. Assessments for naturalistic, story speech in multiple languages require accurate, time-efficient scoring. The current research leverages modern large language models (LLMs) in native English speakers and native speakers of 10 other languages to automate the generation of high-quality, spoken stories and scoring of speech recall in different languages. Participants listened to and freely recalled short stories (in quiet/clear and in babble noise) in their native language. LLM text-embeddings and LLM prompt engineering with semantic similarity analyses to score speech recall revealed sensitivity to known effects of temporal order, primacy/recency, and background noise, and high similarity of recall scores across languages. The work overcomes limitations associated with simple speech materials and testing of closed native-speaker groups because recall data of varying length and details can be mapped across languages with high accuracy. The full automation of speech generation and recall scoring provides an important step towards comprehension assessments of naturalistic speech with clinical applicability.

## Introduction

Many older people live with some form of hearing loss (Cruickshanks et al., 1998; Feder et al., 2015; Goman and Lin, 2016). Peripheral degeneration associated with hearing loss and resulting maladaptive neural plasticity (Auerbach et al., 2014; Salvi et al., 2017; Herrmann and Butler, 2021) impair speech comprehension in noisy situations, such as in crowded restaurants. This often occurs well before pure-tone hearing thresholds worsen sufficiently for hearing loss to be diagnosed (Pichora-Fuller and Levitt, 2012; Pichora-Fuller et al., 2016; Herrmann and Johnsrude, 2020a; Helfer and Jesse, 2021). Persistent speech comprehension challenges increase the risk for social isolation (Shukla et al., 2020; Bott and Saunders, 2021; Motala et al., 2024) and cognitive decline (Lin and Albert, 2014; Chern and Golub, 2019; Livingston et al., 2020). The accurate assessment of speech-comprehension challenges is thus highly relevant for diagnostic and treatment outcomes.

Although individuals seek audiological help mainly because they have difficulties comprehending speech (Portnuff and Bell, 2019; McNeice et al., 2024), only one third of audiologists use speech tests regularly (Portnuff and Bell, 2019; Fitzgerald et al., 2023; Billings et al., 2024). Speech testing add 10-15 minutes to busy ~1-h audiological assessments (Billings et al., 2024), which is time-consuming and costly without automation. Even when speech tests are administered, they are typically available only in a country's dominant language. Speakers of other languages perform worse on speech-in-noise tests despite similar hearing function (van Wijngaarden, 2001; van Wijngaarden et al., 2002; Garcia Lecumberri et al., 2010; Kilman et al., 2015; Phillips et al., 2024). The generation of speech materials in different languages is costly because it requires voice actors with different language backgrounds to record speech. The lack of speech materials and ways to score them in different languages makes assessments less accessible, especially in multi-ethnic countries with a high number of immigrants. Recent advances in AI-based speech synthesizers may open new opportunities (van den Oord et al., 2016; Costa-jussà et al., 2024) because they enable the creation of highly naturalistic-sounding speech materials (Herrmann, 2023) that are increasingly used in research practice (Herrmann, 2024, 2025b; Karunathilake et al., 2025; Pandey and Herrmann, in press) and may have high applicability in clinical settings (Polspoel et al., 2024). Modern synthesized speech can be generated in different languages, potentially reducing the need for costly recordings by voice actors (van den Oord et al., 2016; Polspoel et al., 2024).

Existing speech tests consist of simple words/sentences and verbatim word reports (Killion et al., 2004; Wilson Richard et al., 2007; Wilson et al., 2012; Billings et al., 2024; Polspoel et al., 2024) that poorly resemble the context-rich, story-like speech of typical conversations (Jefferson, 1978; Eisenberg, 1985; Ochs et al., 1992; Mullen and Yi, 1995; Bohanek et al., 2009). Moreover, the mental representations of conversations are not verbatim but reflect the gist integrated across multiple sentences and tens of seconds of speech (Gomulicki, 1956; Mehler, 1963; Mehler and Miller, 1964; Fillenbaum, 1966; Sachs, 1967; Schneider et al., 2002; Martinez, 2024; Raccah et al., 2024; Panela et al., 2025). Verbatim word reports with standard speech tests may fail to capture real-life comprehension challenges because individuals can understand the gist and be highly engaged in a conversation despite missing a significant number of words (Herrmann and Johnsrude, 2020b; Irsik et al., 2022). The focus on verbatim reports of words or short sentences may thus disconnect clinic testing from daily experiences, potentially explaining why hearing aid use is relatively low (Chien and Lin, 2013; Bainbridge and Wallhagen, 2014; Davidson et al., 2021). However, speech tests for naturalistic, story-like speech have thus far not been feasible because they would add ~30 min of manual scoring that audiologists cannot afford. Automating speech generation and speech-comprehension scoring in different languages could advance audiological practice and make assessments of hearing function more accessible.

Recent research suggests that scoring free recall data can be automatized using modern data-science approaches (Heusser et al., 2021; Georgiou et al., 2023; Shen et al., 2023; Martinez, 2024; Raccah et al., 2024). For example, prompt engineering with large-language models (LLMs) (Chen et al., 2024) have been used to score the number of details a person recalls from short passages with a predefined number of details (Georgiou et al., 2023; Martinez, 2024), but it is unclear whether this is scalable to longer more naturalistic materials for which the details are not defined a priori. Another approach is to use either hidden Markov models or, more recently, LLMs to obtain text-embeddings as the basis for analysis (Chandler et al., 2021; Heusser et al., 2021; Shen et al., 2023; Martinez, 2024; Raccah et al., 2024; Herrmann, 2025a). Text-embeddings are high-dimensional vectors that represent the semantic meaning of text pieces (Mikolov et al., 2013; Pennington et al., 2014; Cer et al., 2018; Devlin et al., 2019; Feng et al., 2022; Feuerriegel et al., 2025). For example, the embedding vectors for two semantically related words, such as 'sock' and 'foot', correlate higher than the embedding vectors for two semantically unrelated words, such as 'sock' and 'pool'. Correlations between embedding vectors can thus be used to assess the semantic similarity between segments of the speech materials and a person's recall. The approach enables assessing critical comprehension metrics, including recall magnitude, temporal order recall, and benefits from primacy and recency information in the stimulus materials (Heusser et al., 2021). However, the extent to which it can be used for spoken speech recall across languages is unclear.

The current study expands initial work (Heusser et al., 2021; Shen et al., 2023; Raccah et al., 2024) to explore a completely automated approach to assess speech comprehension of short stories under different speech-clarity conditions (speech in quiet/clear and in 12-talker babble) across 11 different languages. The work uses LLMs to generate stories, translate them into different languages, synthesize auditory stories in different languages, and automatically score speech recall in a listener's own native language (in which they also heard the stories). The approach proposed here proves to be sensitive to known comprehension effects (speech clarity, temporal order, primacy/recency) and generalizes to speakers of different languages. The current work has the potential to open new avenues for an accessible clinical practice that focuses on naturalistic speech listening.

# Methods and materials

## Participants

Two groups of participants took part in the current study. One group included 29 native English speakers or individuals who grew up in English-speaking countries (mostly Canada) and have been speaking English since early childhood (<5 years of age) (mean age: 22.9 years, age range: 19–34 years, 6 male, 21 female, 1 transgender male, 1 non-binary). The other group comprised 26 native speakers of a different language (mean age: 22.9 years, age range: 18–37 years, 7 male, 17 female, 1 genderfluid, 1 preferred not to answer), but who also spoke English sufficiently to understand and comply with task instructions presented in English. Participants of this group were Danish (N=2), French (N=3), German (N=1), Italian (N=1), Polish (N=1), Portuguese (N=2), Russian (N=4), Spanish (N=9), Tagalog (N=2), and Ukrainian (N=1) native speakers. For simplicity, the two groups are referred to as English speakers and non-English speakers, respectively.

All participants reported having normal hearing abilities and no neurological disease. Participants gave written informed consent prior to the experiment and received $20 per hour for their participation. The study was conducted in accordance with the Declaration of Helsinki, the Canadian Tri-Council Policy Statement on Ethical Conduct for Research Involving Humans (TCPS2-2014), and was approved by the Research Ethics Board of the Rotman Research Institute at Baycrest Academy for Research and Education (REB #23-11).

Data from three additional participants were recorded but excluded from analysis. One English speaker reported having a neurological disease. One non-English speaker indicated that their first language is Polish, but they spoke English growing up and felt not very comfortable in Polish anymore. Another non-English speaker indicated that they are fluent in Lithuanian but identified as native English speaker.

## Speech materials

Speech materials were auditory stories of about 2-min duration. The aim of the current study was to fully automate stimulus generation and data analysis approaches to test the feasibility of examining speech comprehension in individuals with different language backgrounds. Hence, stories were generated using generative AI for text generation and AI-based text-to-speech synthesizers (Georgiou et al., 2023; Herrmann, 2024).

In detail, OpenAI's GPT-4o (v2024-05-13; OpenAI et al., 2023) was used to generate seven story texts (Georgiou et al., 2023; Herrmann, 2024), one for practice and six for the main experiment. Each story text focused on a different topic (e.g., a boy stealing and being caught; a women becoming a construction worker). The model was prompted using a zero-shot approach (Chen et al., 2024) with the following input: "Generate a 300 word engaging story about [story topic was inserted here]. The story should not start with "once upon a time" and should not contain quotations, dashes, colons, or semicolons. Generate a related story title.". The number of words in the stories ranged between 314 to 355 (English). Each story text was concatenated into a single paragraph for stimulus generation and analysis.

Auditory stories were generated from the story texts using OpenAI's Whisper (model: 'tts-1-hd'; Radford et al., 2022) through OpenAI's application processing interface (API) using Python code. Whisper is an automatic speech recognition system that was trained on 680,000 hours of multilingual and multitask supervised data collected from the internet (Radford et al., 2022). Whisper supports text-to-speech and speech-to-text applications in different languages. To generate auditory stories in English, each story text was directly fed into the Whisper API. To generate auditory stories in other languages, OpenAI's GPT-4o (v2024-05-13; API; OpenAI et al., 2023) was used to translate each story, using a zero-shot prompt (Chen et al., 2024): "Translate the following text to [language]: ", followed by the story text. The translated text was fed into Whisper. Piloting with colleagues in the lab suggested accurate translation and high auditory quality for several languages. However, piloting also suggested that Asian languages, such as Cantonese, Mandarin, Bengali, and Hindi are more poorly converted to auditory stories using Whisper (despite accurate translations). The current study therefore focused on languages for which we were confident that the audio quality was high. The Whisper voice 'alloy' was used to generate all stories, because this voice is naturalistic sounding and gender neutral. Indeed, individuals experience modern AI speech as very naturalistic and speech perception in quiet and noise is highly similar for AI and human speech (Herrmann, 2023). The 'alloy' voice was used for the synthesis of all stories in different languages, having the advantage that acoustic confounds are minimized that would arise if different voice actors had recorded the stories. Whisper generates audio files at a 24-kHz sampling frequency. Each story was up-sampled to a 44.1 kHz sampling frequency to enable use with the sound-delivery system.

To investigate the sensitivity of the recall scoring approach to different speech listening conditions, each participant was presented with three stories under clear conditions and three stories with added 12-talker babble (Bilger, 1984; Bilger et al., 1984) at a signal-to-noise ratio (SNR) of +2 dB. Twelve-talker babble simulates a crowded restaurant, while not permitting identification of individual words in the masker (Bilger et al., 1984; Mattys et al., 2012). Young normal-hearing listeners correctly hear about 80% of words of continuous speech masked by a twelve-talker babble at +2 dB SNR (Irsik et al., 2022).

For each participant group (English, non-English), half of the participants listened to the stories with the topics "A boy being surprised", "A man going camping for the first time", and "A woman becoming a construction worker" under clear conditions (i.e., in quiet) and to the stories with the topics "'A boy stealing and being caught'", "A woman moving into a shared flat", and "A woman surprising her grandma" in babble. For the other half of the participants, the assignment of speech-clarity conditions (clear, babble) was reversed; counter-balanced as much as possible across languages.

**Experimental procedures**

Participants sat in a comfortable chair inside a single-wall sound booth (Eckel Industries). Experimental procedures were run using Psychtoolbox (v3.0.14) in MATLAB (MathWorks Inc.) on a Lenovo T450s laptop with Microsoft Windows 7. The laptop screen was mirrored to a ViewSonic monitor within the sound booth. In each of the 6 blocks, participants listened to one 2-min story and afterwards freely recalled the story. Participants were instructed to recall as many details as possible. Story order was randomized for each participant. Participants were familiarized with the task procedures in one training block with a separate 2-min story, prior to the 6 experimental blocks. Instructions for all participants were in English and all participants indicated they understood the instructions (they were all proficient English speakers). Critically, stories were played to people in their native language, and participants were instructed to recall the story in their native language. Auditory stories were presented at a comfortable listening level that was fixed across participants (~70 dB SPL) via Sony Dynamic Stereo MDR-7506 headphones and an RME Fireface 400 external sound card. Participants' free recall was recorded via a Shure SM7B microphone.

**Language and speech-quality ratings**

Following the main experimental procedures, participants rated their experience (Likert scale from 1 to 5) with the spoken story materials (exact statements are described in Table 1). Ratings were obtained only for a subset of participants (19 English speakers; 18 non-English speakers), because the procedures were implemented after the study had already started. English and non-English speakers did not differ in their ratings of the overall impression of the stories, the naturalness of the spoken speech, nor the accurateness of the grammar (for all $p > 0.1$; Table 1). Non-English speakers rated speech understanding, the accurateness of the speech's meaning, and vocabulary usage slightly lower than English speakers (for both $p < 0.05$; Table 1), but the rating scores were relatively high for both groups. The most notable difference was that non-English speakers rated the speech pronunciation lower than English speakers and that they more frequently felt the speech was accented (for both p < 0.001; Table 1).

| Statement | English speakers | non-English speakers | Statistic |
|---|---|---|---|
| My overall impression of the stories was good. *('strongly disagree' … 'strong agree')* | 4.15±0.49 | 3.83±0.69 | $t_{35} = 1.617$, p = 0.115 |
| The spoken speech sounded naturalistic. *('strongly disagree' … 'strong agree')* | 3.68±0.98 | 3.05±1.22 | $t_{35} = 1.685$, p = 0.101 |
| I could understand everything (when there was no background noise). *('strongly disagree' … 'strong agree')* | 4.68±0.57 | 4.17±0.76 | **$t_{35} = 2.284$, p = 0.029** |
| The pronunciation of the spoken speech was … *('Poor' … 'Excellent')* | 4.58±0.67 | 3.06±0.91 | **$t_{35} = 5.643$, p < 0.001** |
| The language was grammatically/syntactically accurate. *('Not at all' … 'No mistakes')* | 4.53±0.99 | 4.0±0.94 | $t_{35} = 1.607$, p = 0.117 |
| The meaning of the words and sentences was accurate. *('Not at all' … 'No mistakes')* | 4.68±0.57 | 4.17±0.76 | **$t_{35} = 2.284$, p = 0.029** |
| The vocabulary usage was … *('Very poor vocabulary, inappropriate or incorrect usage' … 'Excellent vocabulary usage, fully appropriate')* | 4.53±0.59 | 3.72±0.87 | **$t_{35} = 3.206$, p = 0.003** |
| The spoken speech sounded accented. *('Yes', 'No')* | p(yes) = 0.052 | p(yes) = 0.889 | **$\chi^2 = 22.76$, p < 0.001** |

**Table 1: Ratings of speech materials. Left column:** Displays the statement which participants (19 English speakers; 18 non-English speakers) rated on a 5-point scale (scale end points are indicated in parentheses) and the statement for which they answered 'yes' or 'no'. Middle columns: Mean and standard deviation for each statement and speaker group (English, non-English). Right column: Test statistics showing t-values (or chi-squared value for the 'yes' vs 'no' question) and p-values. Text indicated in bold showed a significant difference between English and non-English speakers (p < 0.05). Note that the identical descriptive and test statistics for the statements 'The meaning of the words and sentences was accurate' and 'I could understand everything' are not mistakes. The values for individual participants differ between the two statements; only the summary statistics happened to be identical.

The results of the ratings are also reflected in qualitative reports from participants during debriefing. Non-English speakers reported that they liked the materials and could understand the speech, but that occasionally a word was presented in a context in which it would not typically occur (relating to the translation quality), and that the speech sometimes sounded like a person speaking with an English accent (relating to speech synthesis quality). The English accent of OpenAI voices in non-English languages has been noted previously in developer forums (e.g., https://community.openai.com/t/tts-voices-have-a-clear-us-accent/705394), because the model was optimized for English (https://platform.openai.com/docs/guides/text-to-speech).

The group differences in ratings and qualitative reports highlight existing limitations of modern speech synthesis in different languages. Critically, the current study aims to automate the scoring of spoken recall of native speakers of different languages and is thus not directly concerned with the quality of the speech materials. Although differences in speech synthesis quality between groups could in principle influence recall performance, this does not seem to be the case in the current study (as described below). Nevertheless, potential limitations arising from group differences in rating scores will be detailed in the discussion section.

**Preprocessing of recall data**

Verbal recall data were converted to text using OpenAI's Whisper speech-to-text model (whisper-1) using the OpenAI API through Python. As outlined above, Whisper is a multi-language model and can provide transcriptions for speech in different languages. For some transcriptions, punctuation was largely absent, because the model appeared to interpret the audio signal as one stream of thought. OpenAI's GPT-4o (v2024-05-13; temperature = 0; OpenAI et al., 2023) was used to correct punctuation using the following prompt: "The following text has no or minimal punctuation. Please repair missing punctuation where it seems appropriate. Use only periods, commas, and question marks. Avoid all other punctuation. Be moderate and do not overdo it. Do not add, replace, or remove any words from the text. Do not insert line breaks. Add punctuation, but otherwise reproduce the text exactly. This is the text for which punctuation needs to be repaired: ", followed by the transcription. Punctuation correction was carried out for all recall data to avoid manual evaluation. A qualitative assessment suggested that when punctuation was already present, GPT-4o did not add unnecessary punctuation. More generally, the presence or absence of punctuation had little impact on recall analyses. The punctuation-corrected transcriptions in the original language were used as the recall data for analysis. That is, all story and recall texts used for analysis were in the language in which a person heard and recalled the stories. Analyses were also conducted for recall transcriptions translated to English using OpenAI's GPT-4o, but the results were largely comparable and are thus not reported in the current article.

Analysis of recall data was based on and extended different previous approaches that aimed to automate the analysis of free recall data (Heusser et al., 2021; Georgiou et al., 2023; Shen et al., 2023; Raccah et al., 2024; Panela et al., 2025). The approach implemented here uses modern LLMs and enables analyses across languages that were not possible with previously used models. Analyses involve mapping the story texts and recall transcriptions onto high-dimensional numerical vectors (i.e., text-embeddings) to capture the semantic meaning of the texts (Mikolov et al., 2013; Pennington et al., 2014; Cer et al., 2018; Devlin et al., 2019; Radford et al., 2019).

Extensive previous work shows that individuals perceive, encode, and recall dynamic, naturalistic environments as discrete events or segments (Speer et al., 2004; Kurby and Zacks, 2008; Sargent et al., 2013; Kurby and Zacks, 2018; Sasmita and Swallow, 2022). Hence, each story text and recall transcription (both in the language spoken by the participant) was divided into 10 equally-sized segments (based on the number of words in a story) with a length that achieved 20% overlap (Figures 1A and 1B). Segmentation also mitigates potential semantic losses resulting from an analysis that aims to capture the semantic space of the entire story by compressing it into a single embedding vector. The 20% overlap was used to minimally smooth recall data over time and avoid artificial text breaks. Ten segments were selected as a compromise between temporally resolved recall scoring and a high number of short segments that may not capture recall appropriately. The impact of the number of segments was analyzed explicitly as described below.

Each segment of the story texts and recall transcriptions was mapped onto one 768-dimensional embedding vector using the Language-agnostic BERT Sentence Embedding (LaBSE) model (Feng et al., 2022), where BERT refers to Google's Bidirectional Encoder Representations from Transformers (Devlin et al., 2019). LaBSE is a multilingual text-embedding model optimized for over 100 languages and is freely available through the Huggingface platform (https://huggingface.co/sentence-transformers/LaBSE). The model was imported into Python through the sentence-transformers module. Running the LaBSE model for each story segment and transcription segment resulted in 10 embedding vectors for each of the 6 story texts, and 10 embedding vectors for each of the 6 recall transcriptions per participant. The embedding vectors were used to calculate different metrics that assess recall performance.

**Automated scoring of recall data**

For each participant and story, all-to-all Spearman correlations were calculated between the 10 embedding vectors reflecting the 10 story-text segments and the 10 embedding vectors reflecting the 10 recall-text segments (Figure 1C). Spearman correlation rather than Pearson's correlation, cosine similarity, or Euclidean distance was used because it is less sensitive to outliers, if there were any. (Zhelezniak et al., 2019; Herrmann, 2025a; Panela et al., 2025). The correlation values of the resulting 10 story segments by 10 recall segments correlation matrix were Fisher Z transformed (Figure 1D). A high value in the correlation matrix means that the semantic content of a story segment is highly similar to the semantic content of a recall segment. 'Control' correlation matrices were calculated by pairing a specific story separately with the recall of the other five stories from the same participant. Overall, the procedure resulted in one story × recall matrix and five control story × recall matrices for each participant and each of the six stories (three clear stories; three stories in background babble).

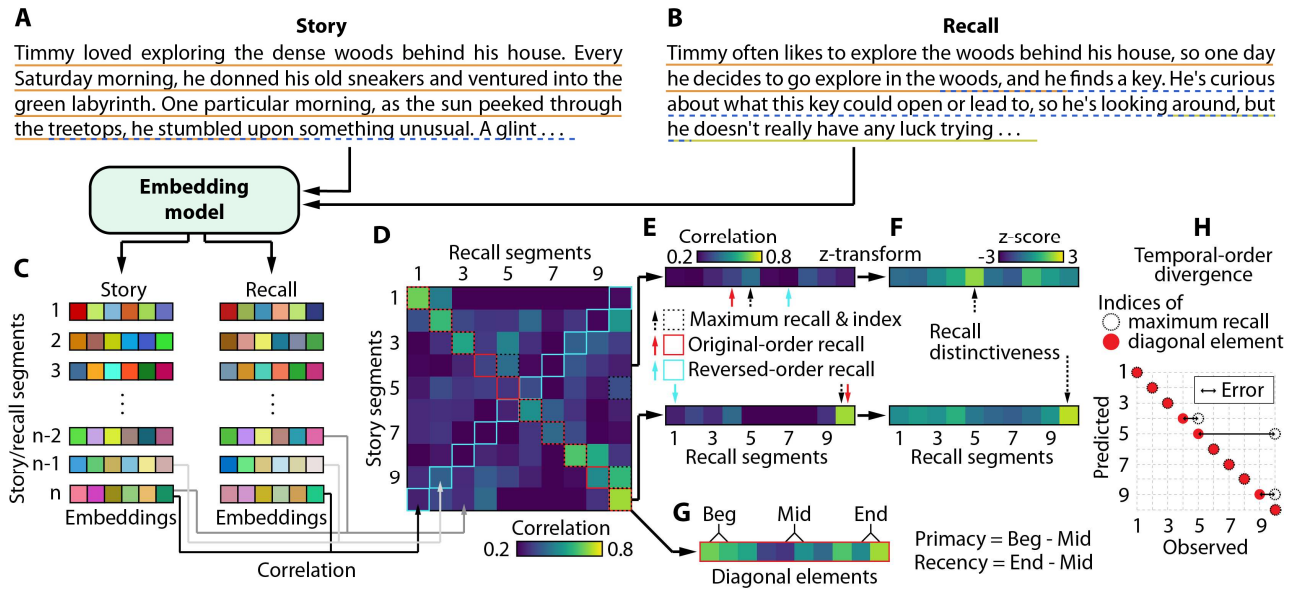

**Figure 1: Schematic and sample data displaying the calculation of recall metrics. A and B:** Sample story and recall texts from one English speaker. The colored lines below the text indicate the story and recall segments, dividing both the story and the recall into 10 equally-sized segments (based on the number of words; separately for story and recall texts) with 20% overlap. **C:** Story and recall segments are fed into a text-embedding model (LaBSE), resulting in one high-dimensional, text-embedding vector for each story and recall segment; here schematically shown, with colors symbolizing the magnitude of embedding dimensions. **D:** The story × recall matrix shows the Spearman correlations (transformed to Fisher's Z correlations) between the embedding vectors of the story segments and the embedding vectors of the recall segments. The colored and line-dashed square boxes within the matrix refer indicate the different recall metrics specified in panel E. **E:** Several recall scores can be calculated from the story × recall matrix. The 'maximum recall score' for a story segment is calculated as the maximum correlation in the corresponding row of the matrix. The index corresponding to the maximum correlation can be used to calculate a 'temporal-order divergence score' represented in panel H. The 'original-order recall scores' are the correlation values along the diagonal of the story × recall matrix. This score is high if a participant recalled the story in the original temporal order. The 'reversed-order recall scores' are the correlation values along the reversed diagonal of the story × recall matrix. These values can be used as a control relative to the 'original-order recall scores'. **F:** The 'recall distinctiveness' is the maximum value of a row of the story × recall matrix after standardizing the correlation values within the row. A high recall distinctiveness score indicates that a story segment correlated strongly with one specific recall segment, rather than being similarly related to multiple segments. **G:** Calculation of recall scores for analysis of primacy and recency effects by averaging the first two diagonal correlation values (primacy) and the last two values (recency) and comparing them to the average across the two values in the middle. **H:** The index corresponding to the recall segment that showed the highest correlation with a story segment

('maximum recall score') can be used to calculate the 'temporal-order divergence score' as the root mean squared error between the predicted temporal order (1, 2, 3, ..., n-1, n) and the recalled temporal order. The outlined approach in panels A to H can also be used to calculate the similarity of participants' recall by calculating participant recall × participant recall matrices.

Specific recall metrics were derived from these story × recall correlation matrices (metrics for the five control matrices were subsequently averaged). In detail, for each story segment (i.e., row of the story × recall correlation matrix), the 'maximum recall score' was taken as the maximum Fisher's Z correlation across the 10 recall segments (Figure 1E). The result was one maximum recall score for each story segment. Scores were averaged across the 10 story segments to obtain one maximum recall score reflecting the participant's overall recall for the specific story. A 'recall-distinctiveness score' was calculated by z-transforming (standardizing) the Fisher's Z correlation values within a row (i.e., a story segment) of the story × recall correlation matrix and taking the maximum z-score (Figure 1F; Heusser et al., 2021). A high recall distinctiveness score indicates that a story segment correlated strongly with one specific recall segment, whereas a low value indicates it being similarly related across multiple of the 10 recall segments (Heusser et al., 2021). The index corresponding to the recall segment that showed the highest correlation with a story segment (i.e., the index of the maximum recall score) can be used to calculate a 'temporal-order divergence score' (Figure 1H). To this end, for each story segment (i.e., row in the story × recall matrix), the index corresponding to the highest correlation was extracted. The root mean squared error was then calculated between the original temporal order (i.e., 1, 2, 3, ..., n-1, n) and the recalled temporal order as indicated by the extracted indices. The root mean squared error was taken as the 'temporal-order divergence score' (Figure 1H).

Previous research suggests that individuals tend to recall events in the temporal order in which they were experienced (Howard and Kahana, 2002; Chen et al., 2017; Heusser et al., 2021; Zhang et al., 2023). The diagonal elements in the story × recall correlation matrix reflect the degree to which the story segments were recalled in the original temporal order, that is, the order in which the story was played. Hence, for each story, the correlation values along the diagonal were averaged to obtain one 'original-order recall score' for each story (Figure 1D,E). To examine whether participants indeed recalled a story in the original temporal order, the correlation values along the reversed diagonal were averaged to obtain one 'reversed-order recall score' for each story (i.e., assuming that participants recalled the last story segment first and the first segment last) (Figure 1D,E). A higher original-order recall than reversed-order recall score would indicate that participants recalled a story in the temporal order in which the story was heard.

Previous memory research also highlights that information in the beginning (primacy) and at the end (recency) is better recalled than information in the middle of stream of information (Murdock Jr, 1962; Howard and Kahana, 1999; Tan and Ward, 2000; Howard and Kahana, 2002; Zhang et al., 2023). To examine whether primacy plays a role in story recall, the first two scores of the 'original-order recall scores' (i.e., the diagonal elements of the correlation matrix) were averaged and compared to the two averaged scores at the middle. The effect of recency was examined by averaging the last two scores of the diagonal elements and comparing it to the two averaged scores at the middle (Figure 1G).

All recall metrics – maximum recall, original-order recall, reversed-order recall, recall distinctiveness, temporal-order divergence, primacy, recency – were calculated for each participant and story (three clear; three in background babble). Recall metrics were averaged separately across the three clear stories and the three stories in background babble. To assess the degree to which meaningful information about story recall is represented in the story × recall matrices (independent of speech clarity), two repeated-measures analysis of variance (rmANOVA) were calculated, using the within-participant factor Score Type (analysis 1: maximum recall vs. chance maximum recall; analysis 2: original-order recall vs. reverse-order recall; collapsed across speech-clarity levels) and the between-participant factor Speaker (English, Non-English). To assess the impact of background babble on story recall, a rmANOVA was calculated with the within-participant factor Speech Clarity (clear, babble) and the between-participant factor Speaker (English, Non-English), separately calculated for the original-order recall score, recall distinctiveness, temporal-order divergence. To assess primacy and recency effects, the rmANOVA included the additional within-participant factor Time (beginning, middle, end). Post hoc comparison were calculated in case of significant effects, using Holm's methods to correct for multiple comparisons (Holm, 1979). Statistical analyses were conducted in JASP software (JASP, 2024; version 0.19.1.0).

**Intersubject correlation of recall**

The recall metrics described in the previous section reflect how well the semantic content of a participant's recall corresponds to the semantic content of a story. In order to investigate how well story recall in different languages agrees across participants, recall agreement metrics were calculated based on previously established intersubject correlation approaches (ISC; Hasson et al., 2004; Hasson et al., 2008; Hasson et al., 2010; Dmochowski et al., 2012; Nastase et al., 2019; Nguyen et al., 2019; Irsik et al., 2022; Lee and Chen, 2022). Calculations were similar to those outlined in Figure 1, with the exception that correlations were calculated between the recall transcriptions of two participants rather than between the story text and the recall transcription of a participant. Recall transcriptions were maintained in their original languages and were not translated to a common language prior to analysis.

In detail, for each participant, the all-to-all Spearman correlations (Fisher's Z transformation) were calculated between the 10 embedding vectors corresponding to one participant's recall and the 10 embedding vectors corresponding to another participant's recall, leading to one recall × recall matrix (using the transcription of the recall in each participant's native language). For each participant, one correlation matrix was calculated relative to each of the other participants listening to the same story under the same speech clarity condition (clear or babble). For example, 14 English and 14 non-English speakers listened to the story "A boy being surprised" in babble noise. For one English speaker out of these 28 participants, 27 recall × recall correlation matrices were calculated, 13 relating to the other English speakers and 14 relating to the non-English speakers. For each recall × recall correlation matrix, we calculated the mean across the diagonal elements ('original-order score'; Figure 1E) – reflecting how much participants recalled the story in a similar temporal order – and the mean across the reversed-diagonal elements as a control ('reversed-order score'; Figure 1E). The 'temporal-order divergence' was also calculated for each recall × recall correlation matrix to assess the degree to which participants diverged from recalling the story in the same temporal order (Figure 1H). In the specific example from a few sentences above, this led to 13 scores per recall metric associated with the participant's 'own' group and 14 scores of the participant relative to individuals from the 'other' group. These scores were averaged separately for the participant's 'own' and 'other' group, leading to two scores per metric. Note that 'group' refers to individuals defined above as belonging to either the English speakers or the non-English speakers (and not unique languages).

To assess whether the there is significant intersubject correlation of recall (independent from speech clarity), scores were averaged across speech-clarity conditions and a rmANOVA was calculated using the within-participant factors Score Type (diagonal vs reversed diagonal) and Reference Group (own, other), and the between-participant factor Speaker (English, non-English). To assess the impact of background babble, two rmANOVAs were calculated, one for intersubject correlation (diagonal of the recall × recall matrix) and one for temporal-order divergence, with the within-participant factors Speech Clarity (clear, babble) and Reference Group (own, other), and the between-participant factor Speaker (English, Non-English). Post hoc comparison were calculated in the case of significant effects, using Holm's methods to correct for multiple comparisons (Holm, 1979).

**Recall scoring using similarity ratings through prompt engineering**

Recall scoring through correlations between embedding vectors for story text segments and recall text segments have been used most for automated recall scoring (Heusser et al., 2021; Shen et al., 2023; Raccah et al., 2024; sometimes cosine similarity is used instead of correlation), and the current work expands the approach by using LLMs, novel metrics, and, most critical, scoring across languages. The embedding approach is computationally fast and is sensitive to experimental manipulations (see below). However, correlations between embedding vectors from multi-word text segments can lead to a small dynamic range of values. Moreover, scores are rarely close to zero, even when embedding vectors from unrelated stories are correlated (e.g., 0.2; see below). For clinical purposes, scores should ideally be easily interpretable (meaningful dynamic range) and low recall should lead to a score close to 0.

Previous work has explored prompt engineering with modern LLMs for shorter stories, such that the LLM provides a rating or accuracy score (Georgiou et al., 2023; Martinez, 2024). The current work extends this work to explore whether LLM prompt engineering provides a greater dynamic range of values and a score close to zero for a recall segment that is unrelated to a story segment. The following zero-shot prompt (Chen et al., 2024) was used to obtain a rating score between 0 and 100 for each of the 10 × 10 combinations of story text segments and recall text segments:

> *Act as an expert rater of speech contents. On a scale from 0-100, rate the degree to which a text segment from a story is semantically captured by the text segment from a human who recalled the story. 0 refers to "not very" and 100 refers to "very". If the recall segment mostly does not capture the story segment, give a 0. If the recall segment mostly captures the story segment, give a 100. This is the text segment from the story: ' [story segment is inserted here]'. This is the text segment from the human recall: ' [recall segment is inserted here]'. Please provide only the rating score and nothing else. Use the full range of the scale.*

An LLM rating value for each story segment and recall segment combination was obtained 3 times and then averaged to reduce noise. Scores were divided by 100 to obtain scores akin to proportion correct responses.

The prompt was provided to OpenAI's 'gpt-4o-mini-2024-07-18' model (temperature = 0). The 'mini' GPT model was used instead of GPT-4o (v2024-05-13), because each prompt fed to the model is associated with costs that were about 17 times cheaper for the 'mini' GPT model. Calculating the 10 × 10 matrices and corresponding chance-level matrices (i.e., relating a story segment to a recall segment of a different story) required 10,800 prompt requests, costing about $1.5 USD per participant for the 'gpt-4o-mini-2024-07-18' model (the GPT-4o model would have

led to >$1200 overall). Practical implications of the approach will be discussed in the discussion section.

After obtaining the story × recall matrices, calculation of recall metrics and statistical analyses mirror the procedures described above for the embedding approach (Figure 1).

Intersubject correlation was not calculated for the GPT rating approach due to the high monetary and computational costs (several hours per participant) associated with the prompting approach for intersubject correlation (>30,000 prompt request), but pilot testing suggested similarly meaningful sensitivity to experimental manipulations. Practical solutions for future research will be discussed in the discussion section.

**Evaluation of the effect of segment numbers**

The analyses described above, and the related results presented below are based on the segmentation of story texts and recall transcriptions into 10 segments. To evaluate whether the number of segments by which story texts and recall transcriptions are divided impacts the sensitivity of the automated assessment approach, calculations and analyses were carried out for 6, 10, 14, and 18 segments. Statistical assessments included similar analyses as described above, focusing on speech-clarity contrasts to assess the current approach's sensitivity.

## Results

**Recall accuracy: Embedding approach**

Figure 2A shows the story × recall correlation matrices for English and non-English speakers. Highest values are present along the diagonal of the matrix, indicating that participants recalled the stories in the original temporal order.

Repeated-measures ANOVAs revealed that the maximum-recall score was greater than chance recall score (i.e., maximum-recall score for unrelated stories) and that the original-order score was greater than the reverse-order score (effect of Score Type; for both analyses: $F_{1,53} >$ 300, $p < 0.001$: Figure 2B), showing that there is meaningful information in the correlation matrices and that stories are recalled in the original temporal order. There was no effect of Speaker (for both analyses $p > 0.7$). There was a Score Type × Speaker interaction for the analysis contrasting maximum-recall score to chance recall ($F_{1,53} = 4.612$, $p = 0.036$, $\omega^2 = 0.012$), because the difference between the maximum-recall score and chance level was slightly smaller for English than non-English speakers.

To provide a visual impression of the similarity of the recall scores across languages, the right-hand side of Figure 2B shows individual recall scores (original-order recall) for each non-English speaker and the distribution of scores for English speakers. All scores for non-English speakers fall within the distribution of scores for English speakers, confirming visually the absence of a difference between speakers. The results suggest that data are comparable across languages.

**Figure 2: Story × recall correlation matrices and recall scores (embedding approach). A:** Matrices are shown for clear stories and stories in babble and for the actual recall scores and chance level (recall paired with an unrelated story). **B:** Recall scores for clear speech and speech in babble. Violin plots, boxplots, and the mean across participants (white circle) are shown for the maximum-recall score, chance maximum-recall score (recall paired with unrelated stories), original-order recall score, and reverse-order recall score. The right plot shows original-order recall scores for non-English speakers and the distribution for English speakers, to display the overlap between speaker groups. **C:** Clear speech versus speech in babble contrasts for original-order recall score (left), recall distinctiveness (middle), and temporal-order divergence (right). Violin plots show the difference between clear speech and speech in babble. **D:** Recall scores for the beginning, middle, and end of a story, assessing primacy and recency effects. Eng – English speakers, nEng – non-English speakers, n.s. – not significant, *p < 0.05, #p < 0.1

Figure 2C shows the effects of speech clarity (clear, babble) for different recall metrics and participant groups. The rmANOVA for the original-order recall revealed larger recall scores for stories under clear than under babble conditions (effect of Speech Clarity; $F_{1,53} = 10.314$, $p = 0.002$, $\omega^2 = 0.017$; Figure 2C, left). There was no difference between speakers and no interaction (for both $p > 0.4$). There were no effects for recall-distinctiveness and temporal-order divergence (for all $p > 0.1$; Figures 2C, middle and right).

The rmANOVA to investigate primacy and recency effects revealed again higher recall scores for clear stories than stories in babble (effect of Speech Clarity; $F_{1,53} = 13.011$, $p = 6.9 \cdot 10^{-4}$, $\omega^2 = 0.020$; Figure 2D). Recall scores were also greater for the beginning than the middle of a story (primacy: $t_{53} = 8.776$, $p_{Holm} = 2 \cdot 10^{-11}$, $d = 0.737$) and the end than the middle of a story (recency: $t_{53} = 3.836$, $p_{Holm} = 3.4 \cdot 10^{-4}$, $d = 0.227$; effect of Time: $F_{2,106} = 50.622$, $p = 3.7 \cdot 10^{-16}$, $\omega^2 = 0.104$; Figure 2D). No other main effects or interactions were significant (for all $p > 0.15$).

**Intersubject correlation: Embedding approach**

Figure 3A shows the intersubject correlation matrices. The rmANOVA revealed greater scores for diagonal elements than reversed-diagonal elements; effect of Score Type: $F_{1,53} = 1584.941$, $p = 3.6 \cdot 10^{-41}$, $\omega^2 = 0.666$; Figure 3B), showing that participants tend to recall a story in a similar temporal order. Correlations between recalls for speakers of their own were greater than correlations between recalls with speakers from the other group (effect of Reference Group: $F_{1,53} = 37.458$, $p = 1.2 \cdot 10^{-7}$, $\omega^2 = 0.032$). There was no overall difference between English and non-English speakers (effect of Speaker: $F_{1,53} = 0.771$, $p = 0.384$, $\omega^2 < 0.001$), but the Reference Group × Speaker, Score Type × Reference Group, and the Score Type × Reference Group × Speaker interactions were significant (for both $F_{1,53} > 9$, $p < 0.005$, $\omega^2 > 0.0005$). Calculating the Score Type × Reference Group rmANOVA separately for English and non-English speakers revealed an interaction only for non-English speakers ($F_{1,53} = 18.068$, $p = 5.6 \cdot 10^{-4}$, $\omega^2 = 0.014$), but not English speakers ($F_{1,53} = 0.082$, $p = 0.776$, $\omega^2 < 0.001$). For English speakers, both the diagonal and reversed-diagonal scores were greater for the own than the other group (for both $p_{Holm} < 0.001$), whereas for non-English speaker this was only the case for the diagonal ($p_{Holm} = 0.023$), but not reversed-diagonal scores ($p_{Holm} = 0.479$). Critically, English and non-English speakers did not significantly differ for any direct contrasts (for all $p > 0.2$; Figure 3B).

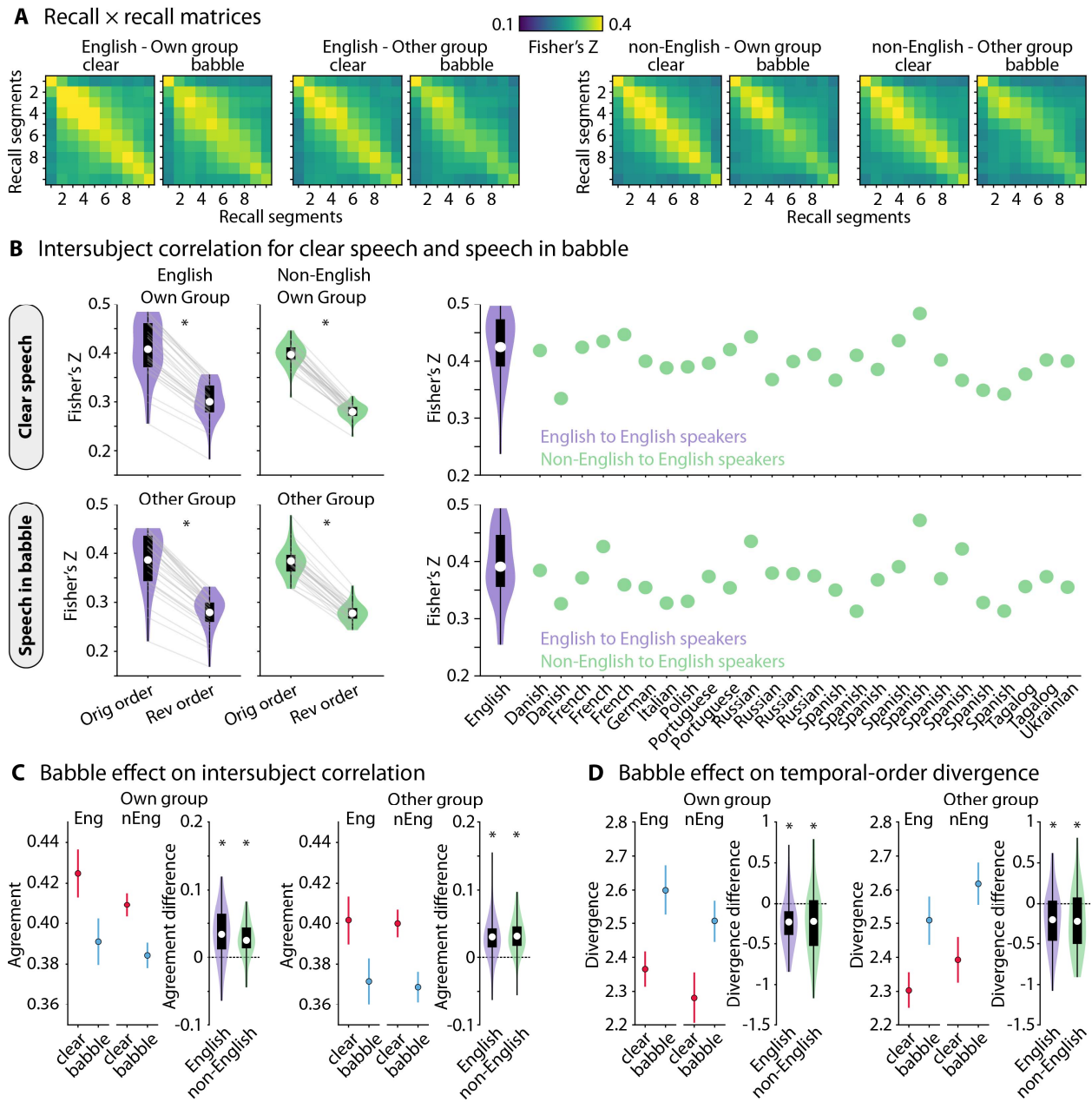

**Figure 3: Recall × recall intersubject correlation matrices and scores (embedding approach). A:** Intersubject correlation matrices are shown for clear stories and stories in babble, for English and non-English speakers, and for correlations among participants of their speaker own group and relative to the other speaker group. **B:** Original order (diagonal of recall × recall matrix) and reversed order scores. Violin plots, boxplots, and the mean across participants (white circle) are shown. Plots on the right show the correlation of recall (original-order scores, i.e., diagonal elements) among English speakers and between non-English and English speakers. **C:** Intersubject correlation (diagonal of recall × recall matrix) and difference between clear speech and speech in babble. **D:** Same as in panel C for temporal-order divergence. Eng – English speakers, nEng – non-English speakers, *p < 0.05.

Figures 3C shows the effect of speech clarity (clear, babble) for intersubject correlation (diagonal elements). Participants' recalls were more similar for clear speech than speech in babble (effect of Speech Clarity: $F_{1,53} = 42.315$, $p = 2.8 \cdot 10^{-8}$, $\omega^2 = 0.084$) and more similar within their own speaker group than relative to the other speaker group (effect of Reference Group: $F_{1,53} = 31.769$, $p = 6.8 \cdot 10^{-7}$, $\omega^2 = 0.030$). There was no difference between English and non-English speakers (effect of Speaker: $F_{1,53} = 0.285$, $p = 0.596$, $\omega^2 < 0.001$), but the Speech Clarity × Reference Group × Speaker interaction was significant ($F_{1,53} = 4.421$, $p = 0.044$, $\omega^2 < 0.001$). Follow-up rmANOVAs revealed a Reference Group × Speaker interaction for clear speech ($F_{1,53} = 7.308$, $p = 0.009$, $\omega^2 = 0.004$), such that the greater the intersubject correlation for the 'own' group compared to the 'other' group was only significant for English speakers ($t_{53} = 6.571$, $p_{Holm} = 1.3 \cdot 10^{-7}$, $d = 0.453$), but not for non-English speakers ($t_{53} = 2.499$, $p_{Holm} = 0.078$, $d = 0.182$). For speech in babble, the Reference Group × Speaker was not significant ($F_{1,53} = 0.269$, $p = 0.606$, $\omega^2 < 0.001$), showing greater 'own' group than 'other' group recall similarity for both English and non-English speakers (for both $p_{Holm} < 0.05$; main effect of Reference Group: $F_{1,53} = 21.406$, $p = 2.4 \cdot 10^{-5}$, $\omega^2 = 0.028$). None of the other interactions were significant (for all $p > 0.1$).

Figures 3D shows the effect of speech clarity (clear, babble) for temporal-order divergence. The rmANOVA revealed a greater divergence for speech in babble than clear speech (effect of Speech Clarity: $F_{1,53} = 20.264$, $p = 3.7 \cdot 10^{-5}$, $\omega^2 = 0.104$). There was no difference in temporal-order divergence between English and non-English speakers (effect of Speaker: $F_{1,53} = 0.006$, $p = 0.937$, $\omega^2 < 0.001$), but Reference Group and Speaker interacted ($F_{1,53} = 13.424$, $p = 5.7 \cdot 10^{-4}$, $\omega^2 = 0.026$). For English speakers, temporal-order divergence relative to speakers of the 'other' group (non-English speakers) was lower than relative to speakers of their 'own' group ($t_{53} = 2.166$, $p_{Holm} = 0.174$, $d = 0.224$), whereas in non-English speakers, temporal-order divergence relative to speakers of their 'own' group was lower than relative to speaker of the 'other' group (English speakers) ($t_{53} = 2.995$, $p_{Holm} = 0.025$, $d = 0.328$). In other words, the recall of both English and non-English speakers was temporally more similar (i.e., lower temporal-order divergence) relative to non-English speakers compared to English speakers. None of the other effects and interactions were significant (for all $p > 0.4$).

**Recall accuracy: GPT ratings through prompt engineering**

Figure 4A shows the story × recall rating matrices for English and non-English speakers for the prompt-engineering analyses. The rmANOVAs showed that the maximum-recall score was greater than chance recall score (i.e., maximum-recall score for unrelated stories) and that the original-order score was greater than the reverse-order score (effect of Score Type; for both analyses: $F_{1,53} > 400$, $p < 0.001$: Figure 4B), showing that stories are recalled more in the original order than in the reversed order. There was no effect of Speaker (for both $p > 0.15$) and no Score

Type × Speaker interaction (for both p > 0.6). Individual data points for non-English speakers relative to the distribution of data for English speakers are shown in Figure 4B, right.

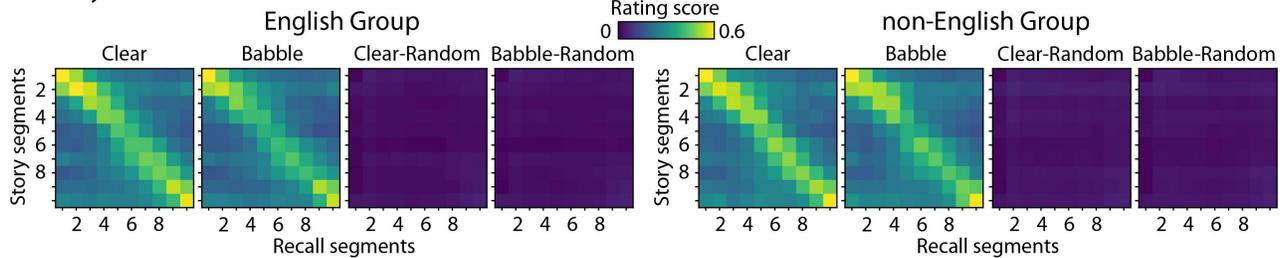

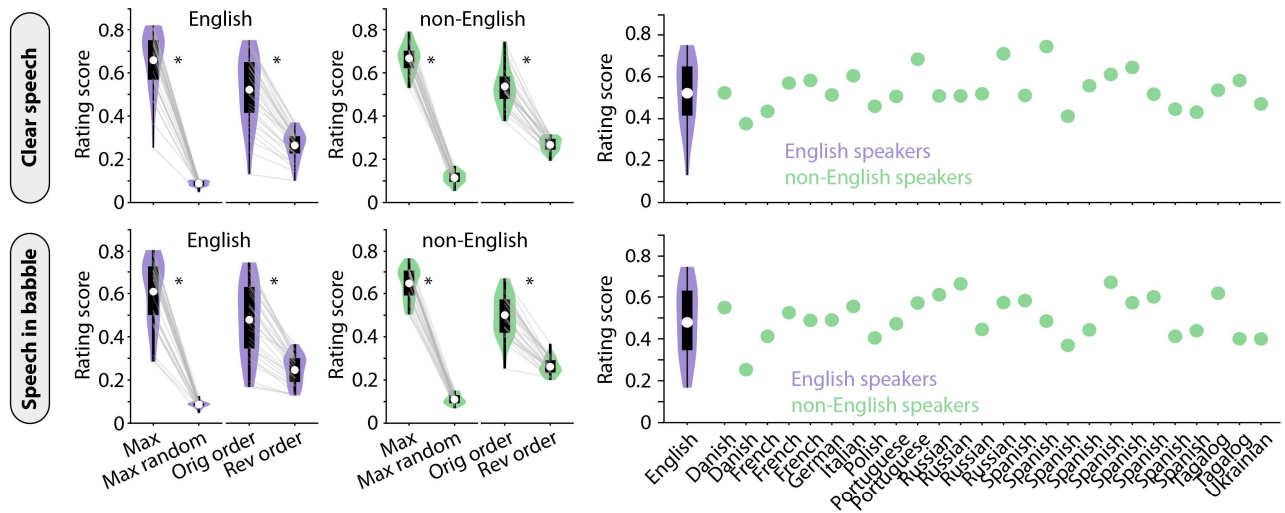

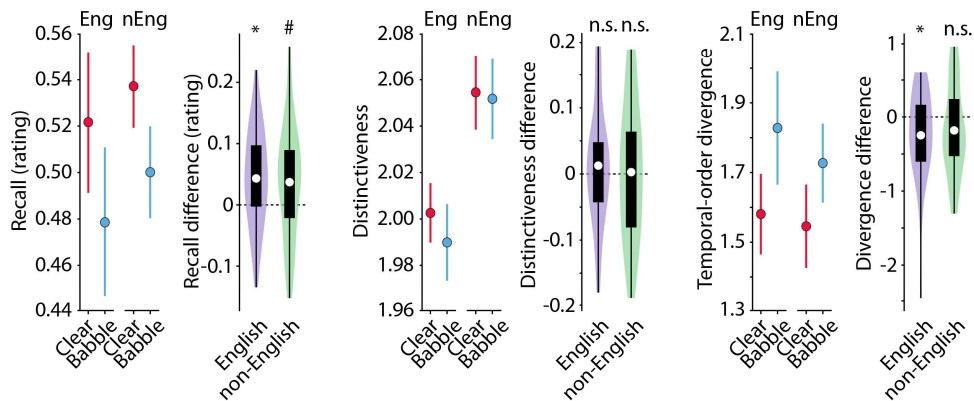
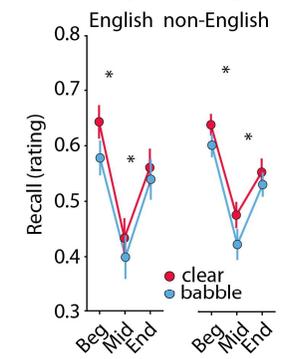

**Figure 4: Story × recall correlation matrices and recall scores (GPT rating approach). A:** Matrices are shown for clear stories and stories in babble and for the actual recall scores and chance level (recall paired with an unrelated story). **B:** Recall scores for clear speech and speech in babble. Violin plots, boxplots, and the mean across participants (white circle) are shown for the maximum-recall score, chance maximum-recall score (recall paired with unrelated stories), original-order recall score, and reverse-order recall score. The right plot shows original-order recall scores for non-English speakers and the distribution for English speakers, to display the overlap between speaker groups. **C:** Clear speech versus speech in babble contrasts for original-order recall score (left), recall distinctiveness (middle),

and temporal-order divergence (right). Violin plots show the difference between clear speech and speech in babble. **D:** Recall scores for the beginning, middle, and end of a story/recall, assessing primacy and recency effects. Eng – English speakers, nEng – non-English speakers, n.s. – not significant, *p < 0.05, #p < 0.1

Figure 4C shows the effects of speech clarity (clear, babble) for different recall metrics and participant groups. The rmANOVA for the original-order recall revealed larger recall scores for stories under clear than under babble conditions (effect of Speech Clarity; $F_{1,53}$ = 11.647, p = 0.001, $\omega^2$ = 0.019; Figure 4C, left). There was no difference between speakers and no interaction (for both p > 0.6). Recall-distinctiveness was greater for non-English than English speakers ($F_{1,53}$ = 10.021, p = 0.003, $\omega^2$ = 0.077; Figure 4C, middle), but there was no effect of Speech Clarity and no interaction (for both > 0.5). Temporal-order divergence was greater for speech in babble than clear speech (effect of Speech Clarity; $F_{1,53}$ = 6.377, p = 0.015, $\omega^2$ = 0.02; Figure 4C, left), whereas there was no effect of Speaker and no interaction (for all p > 0.6; Figures 4C, right).

The rmANOVA to investigate primacy and recency effects revealed again higher recall scores for clear stories than stories in babble (effect of Speech Clarity; $F_{1,53}$ = 8.398, p = 0.005, $\omega^2$ = 0.016; Figure 4D). Recall scores were also greater for the beginning than the middle of a story (primacy: $t_{53}$ = 15.249, $p_{Holm}$ = $1.5 \cdot 10^{-20}$, d = 1.150) and the end than the middle of a story (recency: $t_{53}$ = 10.256, $p_{Holm}$ = $6.9 \cdot 10^{-14}$, d = 0.713; effect of Time: $F_{2,106}$ = 140.834, p = $1.4 \cdot 10^{-30}$, $\omega^2$ = 0.220; Figure 2D). No other main effects or interactions were significant (for all p > 0.15).

**Evaluation of the number of segments**

The analyses reported above were based on the segmentation of story texts and recall transcriptions into 10 segments. To evaluate whether the number of segments impact the sensitivity of the automated recall scoring approach, the speech-clarity effects (clear vs. babble) reported in the previous sections were calculated for 6, 10, 14, and 18 story-text and transcription segments.

For recall scoring using the embedding approach (Figure 5A), scores were greater for clear speech than speech in babble for all segment conditions (for all $F_{1,53}$ > 9, p < 0.005). This speech-clarity effect did not differ between English and non-English speakers (for all $F_{1,53}$ < 0.7, p > 0.4), and there was no overall difference between speaker groups (for all $F_{1,53}$ < 0.5, p > 0.5). This indicates that recall scoring is sensitive to speech-clarity conditions in different languages regardless of whether the story texts and recall transcriptions are divided into 6, 10, 14, or 18 segments. However, the speech-clarity effect (clear minus babble) declined with the increase in the number of segments (linear trend; $t_{53}$ = -2.779, p = 0.008, d = 0.167; Figure 5B).

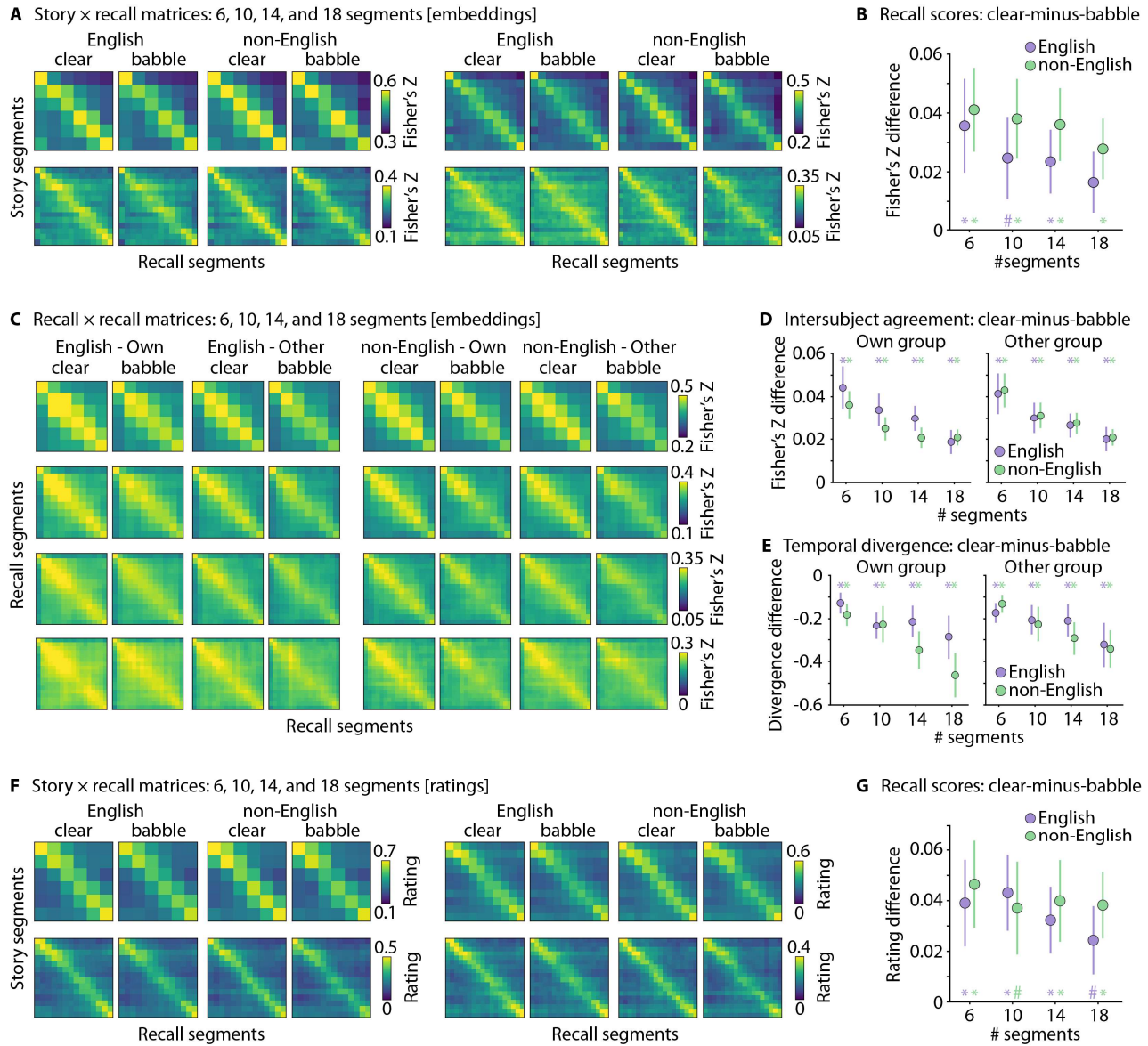

**Figure 5: Results for different segment numbers. A:** Story × recall matrices (from the embedding approach) based on story texts and recall transcriptions divided into 6, 10, 14, or 18 segments (top left, top right, bottom left, and bottom right, respectively). **B:** The effect of background babble (clear minus babble) on recall scores (original-order recall; diagonal elements). The speech-clarity effect decreased for higher segment numbers. The colored * and # indicate p < 0.05 and p < 0.1 significance levels, respectively, for tests against zero. **C:** Recall × recall matrices (from the embedding approach) based on recall transcriptions divided into 6, 10, 14, or 18 segments. **D:** The effect of background babble (clear minus babble) on intersubject agreement (diagonal elements). The speech-clarity effect decreased for higher segment numbers. **E:** The effect of background babble (clear minus babble) on temporal-order divergence scores. The speech-clarity effect increased for higher segment numbers. **F** and **G:** Same as in panels A and B for story × recall matrices from the GPT rating approach though prompt engineering. Colored asterisks indicate a p < 0.05 significance level for tests against zero.

Figure 5C displays the recall × recall matrices (intersubject correlation) for transcriptions divided into 6, 10, 14, and 18 segments. For all segment numbers, intersubject correlation (i.e., the diagonal of recall × recall matrices) was greater for clear speech than speech in babble (for all $F_{1,53} > 35$, $p < 0.001$) and greater among individuals of their own group compared to individuals from the other group (for all $F_{1,53} > 15$, $p < 0.001$; although for 14 and 18 segments, this was greater for English than non-English speakers: Reference Group × Speaker interaction $p < 0.05$; whereas for 10 and 14 segments this difference was specifically greater for English speakers for clear speech: Speech Clarity × Reference Group × Speaker interaction $p < 0.05$). There was no overall difference in intersubject correlation between English and non-English speakers for any of the segment-number conditions (for all $F_{1,53} < 0.9$, $p > 0.35$). The speech-clarity effect (clear minus noise) decreased with the increase in the number of segments (linear trend: $t_{53} = -6.381$, $p = 4.5 \cdot 10^{-8}$, $d = 0.434$; Figure 5D). Temporal-order divergence was greater for speech in babble than clear speech for all segment conditions (for all $F_{1,53} > 20$, $p < 0.001$; Figure 5E). Similar to results reported above (Figure 3D), temporal-order divergence was lower for calculations based on similarity with non-English than English speakers for all segment conditions (Reference Group × Speaker interaction: for all $F_{1,53} > 10$, $p < 0.005$). There was no overall difference between speakers for any of the segment conditions (for all $F_{1,53} < 0.5$, $p > 0.5$). The speech-clarity effect (clear minus noise) increased with the increase in the number of segments (linear trend: $t_{53} = -4.787$, $p = 1.4 \cdot 10^{-5}$, $d = 0.358$; Figure 5E).

For recall scoring using the GPT ratings (Figures 5F and G), scores were greater for clear speech than speech in babble for all segment conditions (for all $F_{1,53} > 10$, $p < 0.003$). The speech-clarity effect did not differ between English and non-English speakers (for all $F_{1,53} < 0.2$, $p > 0.7$), and there was no overall difference between speaker groups (for all $F_{1,53} < 0.5$, $p > 0.4$). This indicates that recall scoring is sensitive to speech-clarity conditions in different languages regardless of whether the story texts and recall transcriptions are divided into 6, 10, 14, or 18 segments. There was no difference in the speech-clarity effect (clear minus babble) for different segment numbers ($F_{1,53} = 1.146$, $p = 0.332$, $\omega^2 < 0.001$; Figure 5G).

Overall, the analyses reported in this section show sensitivity to effects of speech clarity for story texts and recall transcriptions divided into 6, 10, 14, and 18 segments, and no meaningful differences between English and non-English speakers. Nevertheless, the effect of speech clarity for recall scores and intersubject correlation (using the embedding approach) was greatest when fewer segments were used, whereas the speech-clarity effect for temporal-order divergence was greatest when more segments were used, suggesting an intermediate number of segments may be beneficial to maximize sensitivity for different metrics. GPT ratings were similarly sensitive to speech clarity for all segment numbers.

## Discussion

The current study examined whether speech comprehension scoring for naturalistic stories can be automated for listeners with different language backgrounds. The data show that recall scoring using both LLM text-embeddings and LLM prompt engineering is sensitive to relevant speech-comprehension effects, including original temporal-order recall, benefits from primacy and recency information in stories, and reduced recall for speech in background noise. Scoring data for native speakers of various languages showed no meaningful differences, suggesting that LLMs provide a powerful way towards language-independent speech scoring. The current, fully automated approach – from story generation to recall scoring – across languages shows the feasibility of assessing speech comprehension with naturalistic speech materials in speakers with different language backgrounds. The work has the potential to open new avenues for an accessible clinical practice that captures challenges in naturalistic speech listening.

### Large language models for automated comprehension scoring

Manual scoring of verbal or written reports from participants or patients can be extremely time consuming (Levine et al., 2002; Borrie et al., 2019; Bosker, 2021; Martinez, 2024; van Genugten and Schacter, 2024; Herrmann, 2025a). Recent work shows the applicability of LLMs to convert verbal speech to text for speech-recognition use in the speech, language, and hearing sciences (Ballier et al., 2024; Slaney and Fitzgerald, 2024; Zhao et al., 2025). Verbatim word report scoring can also be automatized using LLM text-embeddings (Herrmann, 2025a), but this work has still been limited to words or short sentences.

The scoring of free recall data is substantially more challenging (Levine et al., 2002; Martinez, 2024; van Genugten and Schacter, 2024), especially across languages, because people's mental representations for naturalistic stimuli are not verbatim (Gomulicki, 1956; Mehler, 1963; Mehler and Miller, 1964; Fillenbaum, 1966; Sachs, 1967; Raccah et al., 2024). Each person uses somewhat different words and descriptions when recalling the same story information (Raccah et al., 2024). Scoring in different languages would require a human scorer to be fluent in numerous languages, which makes scalability to a large number of languages not feasible. The current data show that scoring speech recall for speakers with different language backgrounds can be automated using semantic similarity analyses (text-embeddings) and rating scores derived through prompt engineering.

A few other recent works have shown the feasibility of using text-embeddings on segmented recall data for scoring movie comprehension (Heusser et al., 2021; Shen et al., 2023), and recently for story reading (Martinez, 2024; Panela et al., 2025) and listening comprehension (Chandler et al., 2021; Martinez, 2024; Raccah et al., 2024). Some of these works show that scores from text-embeddings correlate well with scores made by humans (Chandler et al., 2021; Martinez, 2024; Panela et al., 2025). The current work extents these previous works by fully automating story generation and scoring in different languages and testing new assessment metrics – that is, temporal-order divergence and intersubject correlation – that show sensitivity to the clarity of speech. Other recent works used LLM prompting to generate short passages with a predefined number of details and LLM prompting to quantify the number of details correctly reported by participants (Georgiou et al., 2023; Martinez, 2024). This work focused on sentence clauses, which are unlikely the unit of mental representations of information encountered in everyday conversations. Nevertheless, scores from prompt engineering showed a high correlation with human raters (Georgiou et al., 2023; Martinez, 2024). Scoring approaches using text-embeddings – despite showing meaningful comprehension sensitivity – can have the disadvantage that even unrelated multi-sentence passages can lead to non-zero correlation/similarity scores (Figures 2A and 2B), thereby reducing the dynamic range and interpretability. By combining the general pipeline of story/recall text segmentation and obtaining metrics from story × recall matrices with LLM prompt engineering, the current study demonstrates a wider dynamic range and chance-level scores close to zero (Figure 4B), while also showing sensitivity to relevant speech-comprehension effects across speakers with different language backgrounds. Combining text segmentation with LLM-based speech-similarity ratings may thus provide the most fruitful avenue for research and clinical practice.

**Sensitivity of LLMs to different aspects of speech comprehension**

The current approach is sensitive to speech comprehension as indicated in several ways. Scores were greater than chance (i.e., when recall data were paired with an unrelated story) and scores were higher for recall in the original than reversed temporal order (Figures 2B and 4B), indicating that participants reported story segments in the order in which they heard a story, which is consistent with previous work (Howard and Kahana, 2002; Chen et al., 2017; Heusser et al., 2021; Zhang et al., 2023). Moreover, recall scores were greater for story segments in the beginning and end compared to the middle of a story, showing the well-known primacy and recency recall pattern (Murdock Jr, 1962; Howard and Kahana, 1999; Tan and Ward, 2000; Howard and Kahana, 2002; Zhang et al., 2023). There were no meaningful differences between English speakers and non-English speakers. In fact, the intersubject correlation analyses demonstrate a similar distribution of correlation scores among the group of English speakers and correlations scores of non-English with English speakers. The automated scoring is thus language independent.

The current scoring approach also shows sensitivity to the clarity of speech. Recall scores were lower for speech in the presence of background noise compared to clear speech. A reduction in speech intelligibility due to background noise is well-known (Miller, 1947; Summerfield, 1987; Gordon et al., 2009; McArdle and Wilson Richard, 2009; Mattys et al., 2012; Pichora-Fuller et al., 2016; Helfer and Jesse, 2021; Pandey and Herrmann, in press), and translates to reduced story comprehension in the current study. Moreover, the temporal order in which participants recalled the stories was reduced for speech in background noise, which appeared to be most prominent in the intersubject correlation analyses (Figure 3D). Temporal-order effects cannot be revealed well with traditional speech-perception assessments, because these assessments rely on individual words or short sentences (Gustafsson and Arlinger, 1994; Killion et al., 2004; Wilson Richard et al., 2007; Wilson et al., 2012; Billings et al., 2024; Polspoel et al., 2024). Automatically scoring the comprehension of naturalistic, story speech thus offers additional avenues to understand speech-comprehension challenges in everyday life.

Young normal-hearing listeners correctly hear about 80% of words of continuous speech masked by the level of background babble used in the current study (+2 dB SNR; Irsik et al., 2022). Participants tend to be as engaged in listening to stories under this level of background babble as they are when listening to clear stories (Herrmann and Johnsrude, 2020b; Irsik et al., 2022), suggesting that missing or incorrectly hearing about 20% of words does not impede comprehension success. This could explain why speech clarity did not affect all recall metrics, and why the significance of the speech-clarity effects varied across analysis types (e.g., recall score vs intersubject correlation).

**Limitations and practical considerations**

The current study provides a language-independent approach to automating the scoring of free-recall data of individuals listening to short stories. The approach is sensitive to critical speech-comprehension effects, but there are practical considerations and limitations.

Stories were generated, translated, and synthesized in different languages using modern large language models, but the rating scores by non-English native speakers highlighted two limitations. Translations occasionally included words that would typically not be used in the context in which they were presented. This limitation is consistent with recent advocacy for including cultural competence beyond text-based information into the training of models for machine translations to improve translations (Tenzer et al., 2024; Feuerriegel et al., 2025). Moreover, the OpenAI voice used in the current study (i.e., Alloy) led to speech that was spoken in a somewhat American-English accent, because the model was optimized for English. The voice was selected because it is naturalistic-sounding and controls voice acoustics across languages. Other synthesizers for different languages are available, such as Google's text-to-speech (https://cloud.google.com/text-to-speech/docs/voices). However, the only highly naturalistic voice available is American-English, whereas voices in other languages still rely on older models that result in less naturalistic speech. The OpenAI voice did not work well with Asian or Arabic languages (despite good translations), leading to incomprehensible auditory materials. This will likely not be an issue for long, given the fast-developing AI space. Despite the language and acoustic limitations, participants with different language backgrounds found the stories engaging and there were no meaningful impacts on the automated scoring.

The automated recall scoring involved dividing story and recall texts into segments. This choice was based on previous recall-scoring work (Chen et al., 2017; Heusser et al., 2021; Raccah et al., 2024) and cognitive research showing that individuals perceive, encode, and recall everyday environments as discrete events (Speer et al., 2004; Zacks et al., 2007; Zacks and Swallow, 2007; Richmond et al., 2017; Lee and Chen, 2022; Sasmita and Swallow, 2022; Pitts et al., 2023). Segmentation was also used to avoid semantic dilution. The current work explored dividing the story and recall texts into different number of segments, showing – for the text-embedding approach – that maximizing the sensitivity to speech-clarity effects may benefit from an intermediate number of segments. If overall recall accuracy is the main focus, fewer segments may be beneficial, whereas sensitivity to effects on temporal-order divergence may be greater for a higher number of segments. The latter may be the case, because more segments provide a higher temporal resolution. For LLM prompting, the number of segments appeared to matter less. Eventually, the choice for the number of segments may also depend on the duration of the story, with longer stories possibly requiring more segments.

The current automated scoring approach leveraged the LaBSE language-independent text-embedding model (Language-agnostic BERT Sentence Embedding; Devlin et al., 2019; Feng et al., 2022). Other multi-language models, such as USE-CMLM-Multilingual (Yang et al., 2020), were also explored, but initial evaluations suggested poorer performance. Initial examinations further suggested that the translation of story and recall texts to English using OpenAI's GPT-4o and then using English-based models or LaBSE lead to qualitative similar results as the ones reported here. Moreover, scoring through LLM prompt engineering did not require translation. It was sufficient to use English instructions with text pieces in other languages. Whether scoring performance could be improved somewhat by using non-English prompts for non-English recall scoring may be worth exploring the in the future, but it appears that state-of-the art models simply map speech across different languages.

Using large language models can come with computational and direct financial costs. OpenAI's LLMs were used in the current study to generate stories, translate speech, and synthesize speech. The costs associated with these tasks involved less than two dollars and were performed within a few minutes for all participants (using a Dell XPS14 laptop). The LaBSE text-embedding model is freely available, and the calculation of the correlation matrices took less than a minute per participant (calculations of intersubject correlations took a few minutes per participant). The LLM zero-shot prompt engineering approach was computationally and financially more expensive even though GPT-4o mini was used. Fees across the calculations conducted were about $80 USD (including chance level) and calculations took about 15 min per participant. Nevertheless, GPT prompt engineering showed somewhat better performance than the embedding approach, such that scores had a greater dynamic range and chance-level scores were near zero. To minimize costs in future work, only the diagonal of the story × recall matrix could be calculated, reducing the number of computations 10-fold (for a 10 × 10 matrix), and calculations for chance level matrices could be omitted. Fees could be avoided by using open-access models like LLaMA 3.1, but this may come with reduced accuracy (LLaMA performs more poorly than GPT; e.g., Panela et al., 2025) and may require higher computational capacity in-house (e.g., computers with GPUs). Other new models such as DeepSeek v3 (https://www.deepseek.com/) may offer a 10 times lower price at the same performance level. Considerations of data sharing, ethical use, and carbon footprint may also determine the model choice (Luccioni et al., 2023; Rillig et al., 2023). The AI landscape is evolving fast, but the current approach outlines the general architecture to automated recall scoring across languages that could likely be used with any future model.

## Conclusions

The current study investigated the feasibility of a fully automated approach to story generation, speech synthesis, and speech-recall scoring across different languages. A group of English speakers and a group of native speakers of a non-English language (10 non-English languages) listened and recalled short stories, both in their native language. The results showed that scoring recall with large language models (LLM) text-embeddings and prompt engineering is sensitive to known speech-comprehension effects, including temporal-order recall, primacy and recency benefits, and sensitivity to reductions in speech clarity. Critically, there were no meaningful differences between English and non-English speakers, demonstrating automated, language-independent scoring of speech recall. The current approach may open fruitful avenues for clinical assessments with naturalistic speech.

## Acknowledgements

I thank Lisa D'Souza for her help with data collection and Ryan Panela for his comments on an earlier vision of the manuscript. The research was supported by the Canada Research Chair program (CRC-2023-00383), the Natural Sciences and Engineering Research Council of Canada (NSERC Discovery Grant: RGPIN-2021-02602), and the Data Sciences Institute at the University of Toronto.

## Author Contributions

**Björn Herrmann:** Conceptualization, methodology, formal analysis, investigation, data curation, writing - original draft, writing - review and editing, visualization, supervision, project administration, funding acquisition.

## Statements and Declarations

The author has no conflicts or competing interests.

## Data Availability

Data are available at https://osf.io/ (upon publication). Three participants declined to share their data publicly (we employ separate consent forms for study participation and data sharing in line with Canadian Tri-Council Policies for Ethical Conduct for Research Involving Humans – TCPS 2 from 2022). Their data are thus not made available.